%% file: main.tex
\documentclass[conference]{IEEEtran}
\IEEEoverridecommandlockouts

\usepackage{cite}
\usepackage{amsmath,amssymb,amsfonts}
\usepackage{algorithmic}
\usepackage{graphicx}
\usepackage{subfigure}
\usepackage{makecell}
\usepackage{flushend}
\usepackage{textcomp}
\usepackage{xcolor}
\usepackage{comment}
\def\BibTeX{{\rm B\kern-.05em{\sc i\kern-.025em b}\kern-.08em
    T\kern-.1667em\lower.7ex\hbox{E}\kern-.125emX}}
\begin{document}
\onecolumn
\title{Prediction Is All MoE Needs: Expert Load Distribution Goes from Fluctuating to Stabilizing}

\author{
Peizhuang Cong, Aomufei Yuan, Shimao Chen, Yuxuan Tian, Bowen Ye, Yilun Yao, Tong Yang
\\
Peking University
\thanks{All authors contribute equally to this paper.}
\thanks{Tong Yang (yangtong@pku.edu.cn) is the corresponding author.}
}

\maketitle

\begin{abstract}
MoE facilitates the development of large models by making the computational complexity of the model no longer scale linearly with increasing parameters. The learning sparse gating network selects a set of experts for each token to be processed; however, this may lead to differences in the number of tokens processed by each expert over several successive iterations, i.e., the expert load fluctuations, which reduces computational parallelization and resource utilization. 

To this end, we traced and analyzed loads of each expert in the training iterations for several large language models in this work, and defined the transient state with ``obvious load fluctuation" and the stable state with ``temporal locality". Moreover, given the characteristics of these two states and the computational overhead, we deployed three classical prediction algorithms that achieve accurate expert load prediction results. For the GPT3 350M model, the average error rates for predicting the expert load proportion over the next 1,000 and 2,000 steps are approximately $1.3\%$ and $1.8\%$, respectively. This work can provide valuable guidance for expert placement or resource allocation for MoE model training. Based on this work, we will propose an expert placement scheme for transient and stable states in our future work.
\end{abstract}


\input{introduction}
\input{relatedwork}

\input{motivation}
\input{statistic_analyses}
\input{evaluation}
\input{conclusion}

\subsection*{Acknowledgments}
This work was supported by the Beijing Natural Science Foundation (Grant No. QY24044). The authors gratefully acknowledge this
financial support, which helped facilitate the research activities
and experimental investigations presented in this paper.

\bibliographystyle{IEEEtran_ref}
\bibliography{reference}
\input{appendix}
\end{document}

%% file: introduction.tex
\section{Introduction}\label{sec:introduction}
In recent years, the magnitude of deep neural networks has scaled up sharply, especially with the emergence of Large Language Models (LLMs), whose parameter quantities have exceeded the trillions level. Although the expansion of parameters enhances the model's ability to handle complex tasks, it also results in enormous computational resources consumption. To alleviate this issue, some LLMs employed the sparse Mixture of Experts (MoE) technique, an architecture that assigns input to some expert networks and integrates their outputs as final output. By employing gating networks to select a subset of experts, sparse MoE LLMs can achieve comparable performance to the dense models while activating fewer parameters, thereby being able to scale model size without proportionally scaling up computational requirements.

Unfortunately, due to the dynamic nature of expert activation requires holding all experts' parameters, MoE LLMs suffer from high memory usage. When allocating GPU resources equally to all experts, it can result in unnecessary wastage of resources due to varying loads of experts (i.e., the volume of processing tokens). Then, some studies tried to add a regularization term to the loss function to balance the load distribution of experts. However, some datasets may be inherently biased towards expert activations, and thus, the accuracy of the model may be affected if the gate networks are over-interfered. From another perspective, the ideal way is to allocate GPU resources to experts based on their individual activity level. The more loaded experts occupy more resources and vice versa, which can guarantee model training efficiency with minimal resources. Nevertheless, it is not easy to make certain of the load of experts during the model training.

In existing researches, some works leverage simple prediction to adjust resources for experts, such as the moving average algorithm (i.e., using the average of historical data as the prediction result directly)\cite{nie2023flexmoe,wang2023prophet}. In fact, such an approach is not always practical. According to our observation and analysis of the expert load data traced from model training, the load of experts has the following characteristics: (a) the load distribution of experts in a MoE layer tends to stabilize gradually as the training iterates, but there are prominent fluctuations at the early stage of training; (b) the load fluctuations of experts in different layers are different, and the fluctuation of load proportion of experts in the shallow layer is more noticeable than deep layers. As aforementioned, predictions can hardly provide valid indications for expert placement or resource allocation in the fluctuation phase. Therefore, it becomes a critical knot when the load proportion of experts starts to convert from fluctuating to relatively stable during the training process. 

To better understand and analyze the expert load during model training, in this work, we conducted extensive training of MoE models under various scenarios (including different cases in terms of model architecture, parameters scale, hyper-parameters, routing strategy, load-balancing loss, dataset, etc.), and traced the load distributions of the experts in all these cases. Based on the in-depth analyses of traced data, we provided insights into the characteristics of the experts load and define transient state and stable state for model training. Moreover, given the computational overhead and the features of both states, we deployed three classical prediction algorithms that can achieve high-precision prediction of expert load distribution. 

In summary, the main contributions of this work are as follows:
\begin{itemize}
    \item We reveal  both the transient and stable states of the expert load during the training process of MoE models.
    \item We conduct extensive experiments to analyze the changing characteristics of the expert load and verify the defined transient and stable states.
    \item We achieve high-precision expert load prediction by using three classical prediction algorithms given the features of transient and stable states.
\end{itemize}

%% file: relatedwork.tex
\section{Preliminary and related work}\label{sec:relatedwork}
\subsection{Mixture of experts}
The Mixture of Experts (MoE) model is initially introduced by \cite{jacobs1991adaptive} in 1991, which integrates different networks through supervised learning, where each network is responsible for processing a specific subset of the training examples. In 2017, Google achieved the first sparsely gated MoE architecture by adding MoE between LSTM layers\cite{shazeer2017outrageously}. 
Given the advantages of sparse MoE architecture over traditional dense networks in terms of capacity and computational efficiency, MoE has been widely employed in various models, especially in transformer-based large language models\cite{lepikhin2020gshard,fedus2022switch,riquelme2021scaling,mustafa2022multimodal}. 

For the MoE architecture, the Gating Network plays a key role, which calculates the weights of each expert based on the input data received from the previous layer, and then distributes the input data to the selected experts, i.e., the activated experts, based on the weights and the assigned selecting rules (e.g., Top-K). 
The results processed by all activated experts will be combined according to the corresponding weights by weighted summation or other forms of integration to generate the final output. 

\begin{figure*}[t]
    \begin{center}
    \subfigure[GPT-3 125M layer-2]{
        \label{subfig:gpt3-125m-layer2_load}
        \includegraphics[width=0.45\textwidth]{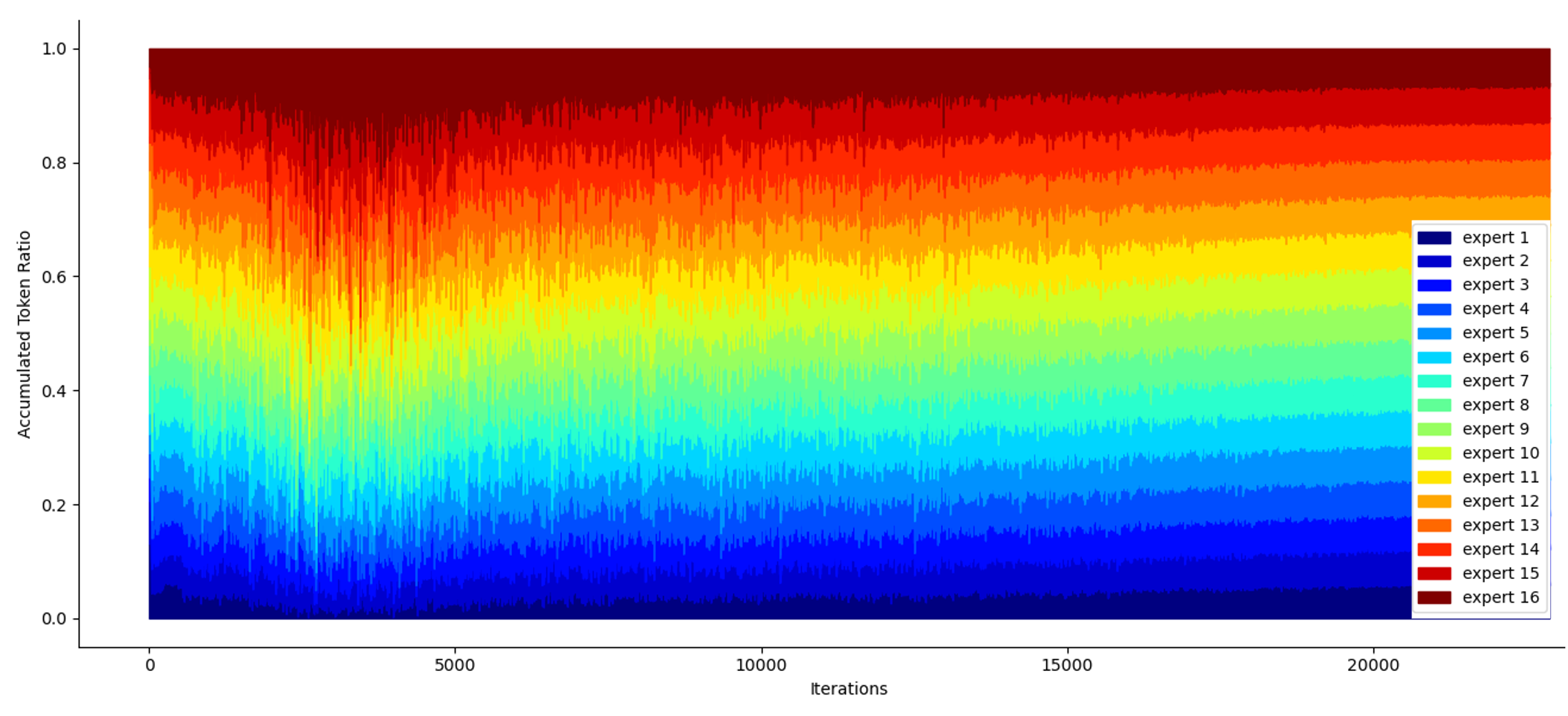}
    }
    \subfigure[GPT-3 125M layer-8]{
        \label{subfig:gpt3-125m-layer8_load}
        \includegraphics[width=0.45\textwidth]{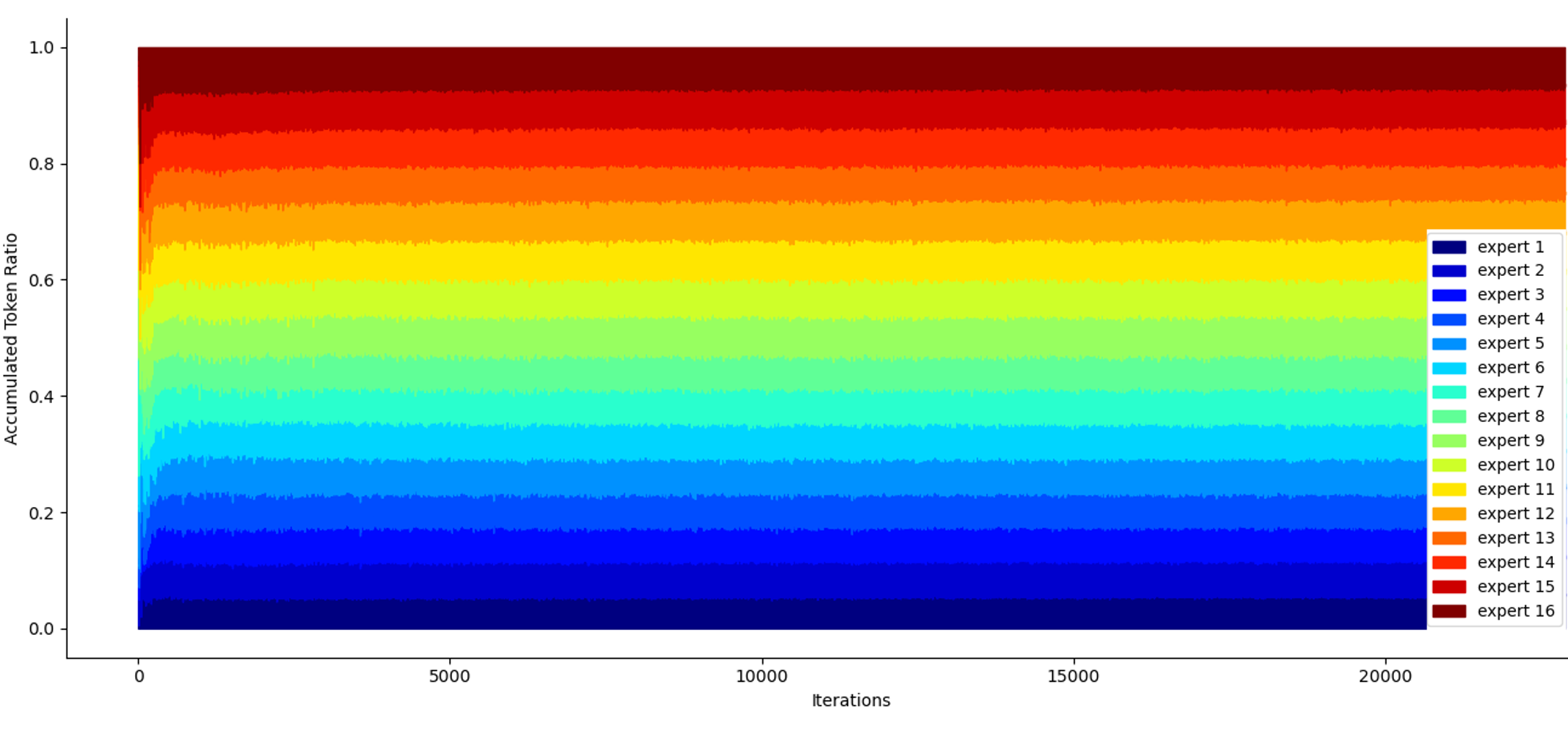}
    }
    \caption{Load proportions of experts in the MoE layer}
    \label{fig:gpt-3-125m-load}
    \end{center}
\end{figure*}

\subsection{Balancing expert load}
The learnability of the gating network makes the probabilities of tokens distributed to experts uncontrollable. In other words, during the training process, it is possible that some experts may receive an excessive number of tokens, while others receive too few. This imbalance in expert load distribution does not align with the intentions of the MoE architecture and can adversely affect computational efficiency. To alleviate this issue, several studies have proposed load-balancing strategies for experts during the model training process. 

\textit{Load balancing loss}: In order to make each set of tokens be distributed to the experts evenly, it is possible to add an auxiliary load balancing loss that usually defined as the sum of the activation entropy of all experts, which ensures that all experts can be trained sufficiently\cite{shazeer2017outrageously,lepikhin2020gshard}.

\textit{Capacity factor}: To further constrain the number of tokens processed by each expert, it is possible to limit the expert capacity, i.e., to stipulate each expert can only process a fixed number of tokens, $CF*(\frac{Num_{tokens}}{Num_{experts}})$. The expert capacity can be adjusted by setting different $CF$ values, and when the experts receive more tokens than the set capacity, these extra tokens  will be passed directly to the next layer through the residual connection\cite{fedus2022switch}.

\textit{Expert-based routing}: In contrast to tokens choose experts, \cite{zhou2022mixture} proposed the way that experts choose tokens. Based on the token-expert affinity scores matrix, which is produced by the dot product of the token embedding and the expert embedding, each expert chooses its corresponding Top-K tokens.

\textit{Hash-based routing}: As different from all of the above, \cite{roller2021hash} and \cite{zhou2023brainformers} replaced the learning gated network with the hash-function, all tokens choose the corresponding experts according to the hash calculation, which can avoid the issue of unevenly balanced load of experts during the training process. 

However, these methods may suffer from the problem that some tokens may be neglected for training the model.

\subsection{Load-based expert placement}
Despite the implementation of various expert load-balancing strategies, imbalances in load distribution still occur in practical scenarios. To ensure the training efficiency of the MoE models, it is possible to adjust the resource allocation for experts or to strategically reallocate experts more effectively. Based on the load of experts, FlexMoE employs a simple but effective heuristic algorithm to dynamically optimize resource placement during model training, enhancing the model training performance \cite{nie2023flexmoe}. Prophet leverages the temporal locality of expert loads to schedule resource allocation operations using a layer-wise, fine-grained strategy\cite{wang2023prophet}.

%% file: motivation.tex
\section{Motivation}\label{sec:motivation}
We traced and investigated the activation frequency of each expert by tokens in each iteration during the training of GPT-3 125M and GPT-3 350M models, which revealed two distinct states of expert load. 
The first state appears in the early iterations of the training, featuring non-regular variations of the load of each expert in successive iterations. We refer to this period as the \textbf{transient state}. With several iterations of training, the second state emerges in which the loads of each expert are similar in adjacent iterations, i.e., showing temporal locality. This phase is referred to as the \textbf{stable state}. 

In the transient state, the intrinsic fluctuation of the loads of the expert makes it hard to get an accurate load proportion by predicting, while it is contrary in the stable state. 
Therefore, accurately distinguishing state transition is crucial for model training. 
This is because load prediction can be leveraged to guide the resource allocation for experts during the stable state. 
In contrast, during the transient state, it is essential to reserve sufficient resources for each expert to cope with load bursts so as to ensure the model training efficiency. 

It is important for resource-efficient training in MoE models to allocate resources flexibly and dynamically according to the load of experts. 
In order to better understand the load state of experts during training process, it is necessary to conduct further investigations and purposive experiments. 
Intuitively, the factors that may affect the transition of the model training include model structure, model parameter scale, model hyper-parameter setting, expert selecting strategy, load balancing loss function, dataset distribution, etc. In this context, we conducted extensive experiments to investigate how these factors influence the changes in expert load states. 
The experiment results indicate that during the model training, the load of the experts transitions from the transient state to the stable state, where the expert load fluctuates initially but gradually stabilizes, exhibiting locality characteristics. 
On basis of that, we conduct three classic algorithms for expert load predicting.

%% file: statistic_analyses.tex
\section{Methodology}\label{sec:method}
\subsection{Preliminary statistic analyses}\label{sec:statistic}

\begin{figure*}[h]
    \begin{center}
    \subfigure[Experts in layer-6, \textit{w=10}]{
        \label{subfig:gpt3-125m-layer2-w10-variance}
        \includegraphics[width=0.3\textwidth]{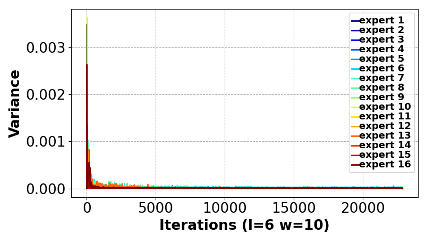}
    }
        \subfigure[Experts in layer-8, \textit{w=10}]{
        \label{subfig:gpt3-125m-layer8-w10-variance}
        \includegraphics[width=0.3\textwidth]{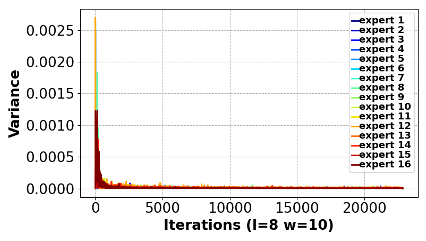}
    }
        \subfigure[Experts in layer-12, \textit{w=10}]{
        \label{subfig:gpt3-125m-layer12-w10-variance}
        \includegraphics[width=0.3\textwidth]{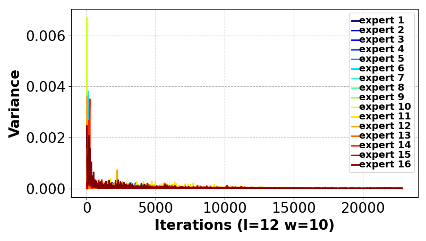}
    }
        \subfigure[Experts in layer-6, \textit{w=100}]{
        \label{subfig:gpt3-125m-layer2-w100-variance}
        \includegraphics[width=0.3\textwidth]{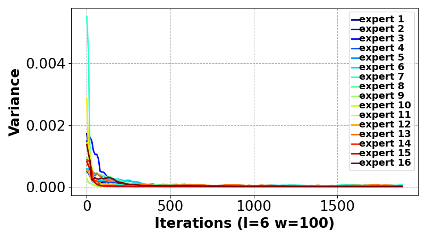}
    }
        \subfigure[Experts in layer-8, \textit{w=100}]{
        \label{subfig:gpt3-125m-layer8-w100-variance}
        \includegraphics[width=0.3\textwidth]{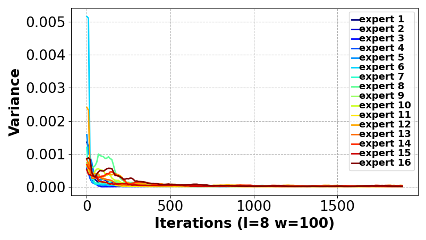}
    }
        \subfigure[Experts in layer-12, \textit{w=100}]{
        \label{subfig:gpt3-125m-layer12-w100-variance}
        \includegraphics[width=0.3\textwidth]{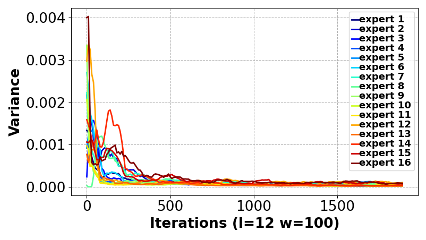}
    }
    \caption{\textsl{Variance} values of experts load proportion of GPT-3 125M (\textit{w=10} and \textit{w=100})}
    \label{fig:gpt-3-125m-w-variance}
    \end{center}
    \begin{center}
    \subfigure[Experts in layer-2]{
        \label{subfig:gpt3-125m-layer2-variance}
        \includegraphics[width=0.3\textwidth]{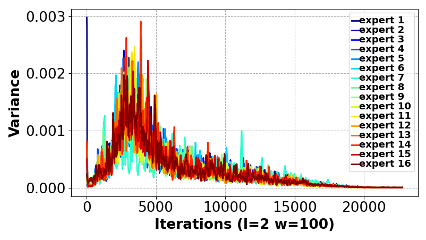}
    }
    \subfigure[Experts in layer-4]{
        \label{subfig:gpt3-125m-layer4-variance}
        \includegraphics[width=0.3\textwidth]{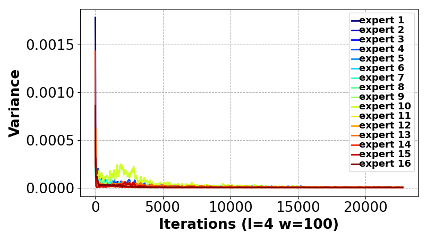}
    }
    \subfigure[Experts in layer-6]{
        \label{subfig:gpt3-125m-layer6-variance}
        \includegraphics[width=0.3\textwidth]{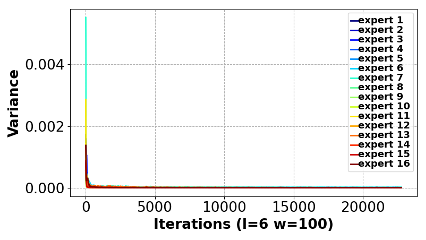}
    }
        \subfigure[Experts in layer-8]{
        \label{subfig:gpt3-125m-layer8-variance}
        \includegraphics[width=0.3\textwidth]{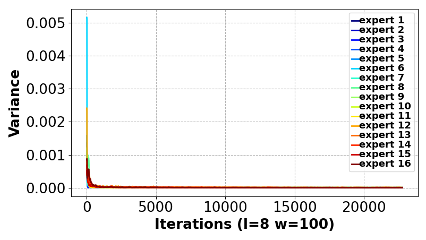}
    }
        \subfigure[Experts in layer-10]{
        \label{subfig:gpt3-125m-layer10-variance}
        \includegraphics[width=0.3\textwidth]{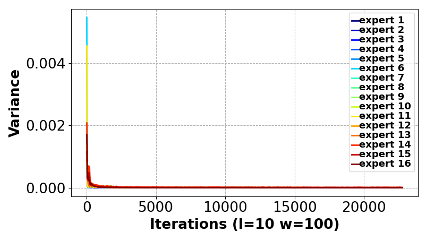}
    }
        \subfigure[Experts in layer-12]{
        \label{subfig:gpt3-125m-layer12-variance}
        \includegraphics[width=0.3\textwidth]{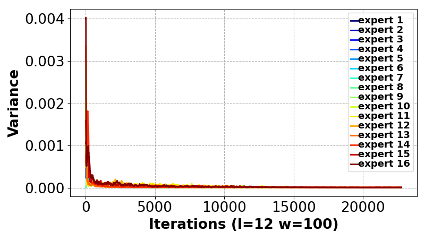}
    }
    \caption{\textsl{Variance} values of experts load proportion of GPT-3 125M (\textit{w=100})}
    \label{fig:gpt-3-125m-variance}
    \end{center}
\end{figure*}

\begin{figure*}[h]
    \begin{center}
    \subfigure[Experts in layer-2]{
        \label{subfig:gpt3-125m-layer2-range}
        \includegraphics[width=0.3\textwidth]{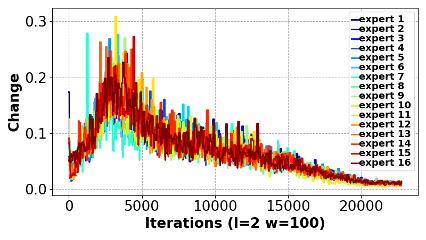}
    }
    \subfigure[Experts in layer-4]{
        \label{subfig:gpt3-125m-layer4-range}
        \includegraphics[width=0.3\textwidth]{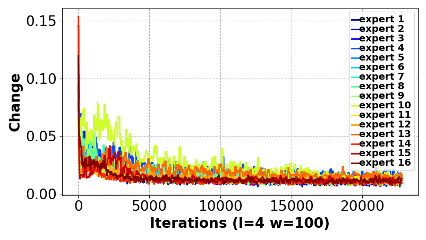}
    }
    \subfigure[Experts in layer-6]{
        \label{subfig:gpt3-125m-layer6-range}
        \includegraphics[width=0.3\textwidth]{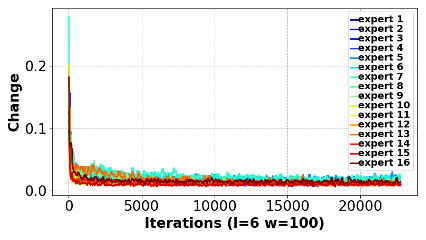}
    }
        \subfigure[Experts in layer-8]{
        \label{subfig:gpt3-125m-layer8-range}
        \includegraphics[width=0.3\textwidth]{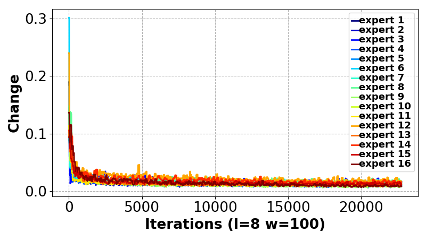}
    }
        \subfigure[Experts in layer-10]{
        \label{subfig:gpt3-125m-layer10-range}
        \includegraphics[width=0.3\textwidth]{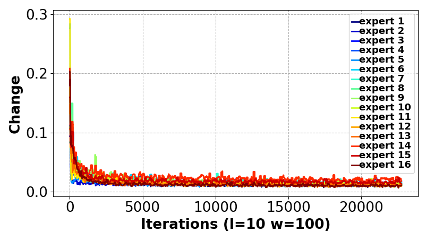}
    }
        \subfigure[Experts in layer-12]{
        \label{subfig:gpt3-125m-layer12-range}
        \includegraphics[width=0.3\textwidth]{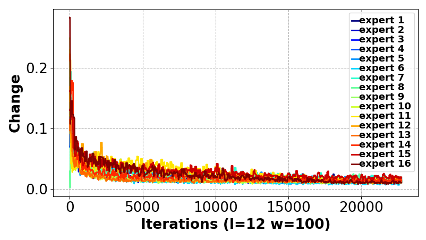}
    }
    \caption{\textsl{Range} values of experts load proportion of GPT-3 125M (\textit{w=100})}
    \label{fig:gpt-3-125m-range}
    \end{center}
\end{figure*}
\begin{figure*}[h]
    \centering
    \subfigure[Accuracy of LSTM-based prediction]{
        \label{fig:gpt3-125m-LSTM-1k}
        \includegraphics[width=0.3\linewidth]{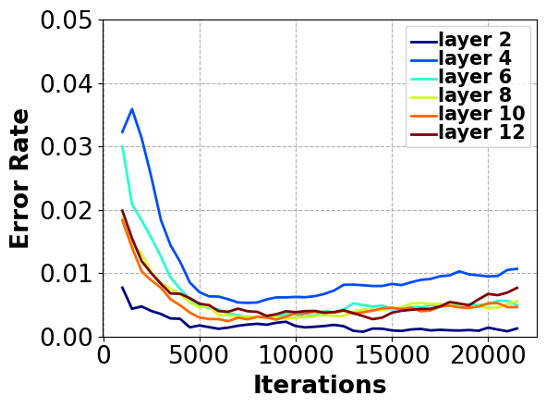}
    }
    \subfigure[Accuracy of ARIMA-based prediction]{
        \label{fig:gpt3-125m-ARIMA-1k}
        \includegraphics[width=0.3\linewidth]{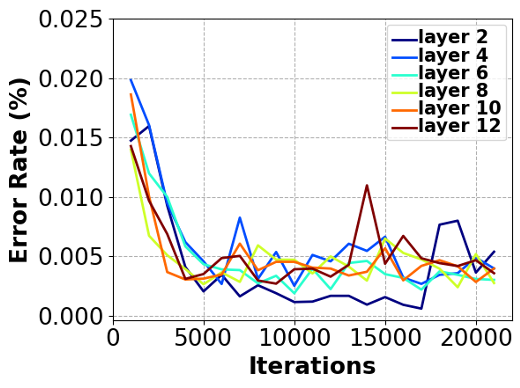}
    }
    \subfigure[Accuracy of LSTM-based prediction]{
        \label{fig:gpt3-125m-AVG-1k}
        \includegraphics[width=0.3\linewidth]{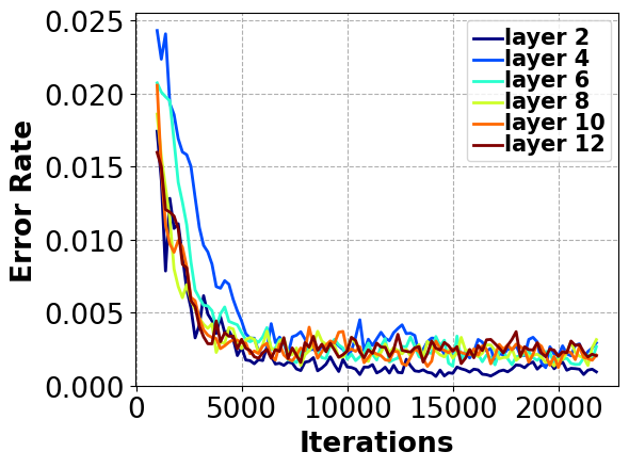}
    }
    \caption{Prediction accuracy for GPT-3 125M}
    \label{fig:gpt3-125m-prediction-accuracy}
\end{figure*}
\begin{table}[h]
    \centering
    \caption{Experiment setup description}
    \label{tab:experiment_setting}
    \begin{tabular}{|c|c|c|} \hline 
         &  \makecell[c]{Experiment setup 1}& Experiment setup 2\\ \hline 
         Model&  GPT-3 Small& GPT-3 Medium\\ \hline 
         Parameters&  125M& 350M\\ \hline 
         Layers (MoE)&  12 (6)& 24 (12)\\ \hline 
         Num of experts&  16 per layer& 128 per layer\\ \hline 
         Hidden size&  768& 1024\\ \hline 
         Num of attention head& 12&16\\ \hline 
         Global batch size& 256&256\\ \hline 
         GPU& 4*4090&4*A800\\ \hline
    \end{tabular}
\end{table}

We initially conducted a statistical analysis of the load distribution of experts in each MoE layer during the model training process under two experimental settings, as outlined in Table \ref{tab:experiment_setting}. Taking the results from \textit{Experiment 1} as an example, as illustrated in Fig. \ref{fig:gpt-3-125m-load}, it is evident that the load characteristics of the experts exhibit the following features: 
\begin{itemize}
    \item In the temporal dimension, the load distribution of experts fluctuates in the early training iterations, and gradually exhibits locality, with minor variations in load distribution between adjacent training iterations;
    \item In the spatial dimension, the first MoE layer (Layer-2) experiences significant fluctuations, while the load distribution of other MoE layers is relatively stable;
    \item As training iterates, the expert load of each MoE layer will convert from transient to stable, i.e., the expert load stabilized gradually.
\end{itemize}

To investigate the state of expert load distribution more intuitively, we quantified the magnitude of changes of each expert load in successive iterations, i.e., we calculate the \textit{variance} ($\frac{1}{Size_{win}}\sum(x_i-\hat{x})$) and \textit{range} ($x_{max}-x_{min}$) of each expert load under different sizes of sliding window. 

Firstly, in the \textit{Experimental setup 1} scenario, we set the sliding window sizes to 10 and 100, and respectively calculate the \textit{variance} of the load proportion of each expert in sliding windows. The results are shown in Fig. \ref{fig:gpt-3-125m-w-variance}. 

Then, in this setup with sliding window size of 100, the results of the \textit{variance} in the sliding window of the expert load proportion for all MoE layers are shown in Fig. \ref{fig:gpt-3-125m-variance}. Moreover, we calculated the \textit{range} values of the each expert in the sliding window with the above setup. The results are shown in Fig. \ref{fig:gpt-3-125m-range}. 

Similarly, we calculate the \textit{variance} and \textit{range} values of each expert load proportion in sliding windows for \textit{Experiment Setup 2}. The detail results are shown in the \texttt{Appendix} part. 

Both the statistical results of \textit{Experiment Setup 1} and \textit{2} demonstrate that each expert successively experiences the transient state and stable state during the model training, i.e., the load proportions of the experts in each MoE layer fluctuates from the beginning of training and stabilizes gradually with iterations of training. Taking into account the inherent characteristics of expert load and the computational efficiency, LSTM-based, ARIMA-based, and sliding window average-based algorithms were selected for the prediction of expert load in this work. The details of these methodologies will be delineated in the next subsection.

\input{methodology}

%% file: methodology.tex
\subsection{Prediction algorithms}\label{sec:methodology}
Based on the statistical characteristics of the expert load distribution, we tried Long Short-Term Memory (LSTM)-based, Auto Regressive Integrated Moving Average (ARIMA)-based, and Sliding Window Average (SW\_Avg)-based algorithms to predict expert load in this work. 
In other words, these algorithms are intended to predict the future load distribution of each expert in the training process in accordance with historical data. 
Given that the total number of tokens $N_{token}$ processed in an iteration by each MoE layer is fixed, it suffices to predict the load proportion of each expert within the MoE layers. 

\begin{itemize}
    \item \textbf{LSTM-based}: The input of the model is $[n^1_1, n^1_2, ..., n^2_1, n^2_2, ..., n^m_e]$, wherein $n^i_j$ represents the historical data of the load proportion of the $j$-th expert in the $i$-th MoE layer, and the output of the model is the load proportion value of all experts in the next $k$ iterations.
    \item \textbf{ARIMA-based}: The ARIMA model is a classic method for time series analysis and forecasting. The ARIMA model is typically denoted as $ARIMA(p, d, q)$, where $p$ is the order of the auto regressive part, $d$ is the degree of first difference involved, and $q$ is the order of the moving average part. The general form of the ARIMA model can be written as: $(1 - \sum_{i=1}^p \phi_i L^i)(1 - L)^d X_t = (1 + \sum_{j=1}^q \theta_j L^j) \epsilon_t$, where $X_t$ represents the time series data, $\phi_i$ is the parameter of the auto regressive part of the model, $\theta_j$ is the parameter of the moving average part, $L$ is the lag operator, such that $L^k X_t = X_{t-k}$, $\epsilon_t$ is the error terms, assumed to be white noise, $d$ is the number of non-seasonal differences needed for stationarity.
    
    The components of the ARIMA model are explained as follows: \textsc{Auto Regressive (AR)} part, $\phi(L) = 1 - \sum_{i=1}^p \phi_i L^i$; \textsc{Integrated (I)}, $(1 - L)^d X_t$; \textsc{Moving Average (MA)} part, $\theta(L) = 1 + \sum_{j=1}^q \theta_j L^j$. The goal of the ARIMA model is to find a model that best fits the time series data by minimizing forecast errors, allowing for accurate future predictions. 
    
    In this prediction algorithm, the tests based on expert load proportion history data for stationarity and seasonality are performed to select appropriate $p$, $d$, and $q$ parameters.

    \item \textbf{SW\_Avg-based}: A straightforward way, taking the arithmetic mean of the data of the load proportion in the historical multiple iterations as the predicted value for the next iteration, and predicting the load of the expert in the future through $k$ rounds of calculation by the means of sliding. This manner exhibits extremely high performance in calculation efficiency, and is also hardware-friendly. Moreover, the experimental results also indicate that this method is simple but effective, and the specific results will be presented in \S \ref{sec:evaluation}.
\end{itemize}

%% file: evaluation.tex
\begin{figure}[t]
    \centering
        \subfigure[Error rate]{
        \label{fig:gpt3-350m-LSTM-1k}
        \includegraphics[width=0.35\linewidth]{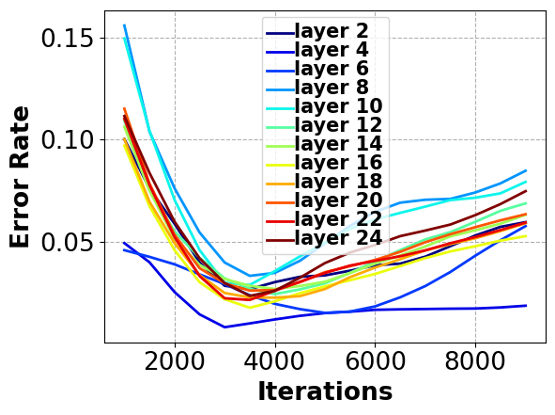}
    }
    \subfigure[Error rate per 1,000 iterations]{
        \label{fig:gpt3-350m-LSTM-1k-bar}
        \includegraphics[width=0.35\linewidth]{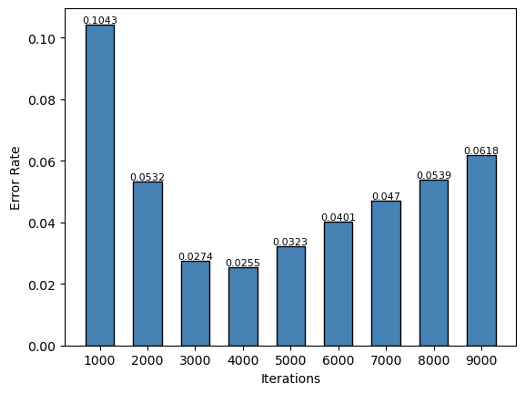}
    }
    \caption{Accuracy of LSTM-based prediction for GPT-3 350M}
    \label{fig:gpt3-350m-LSTM}
\end{figure}
\begin{figure}[t]
    \centering
        \subfigure[Error rate]{
        \label{fig:gpt3-350m-ARIMA-1k}
        \includegraphics[width=0.35\linewidth]{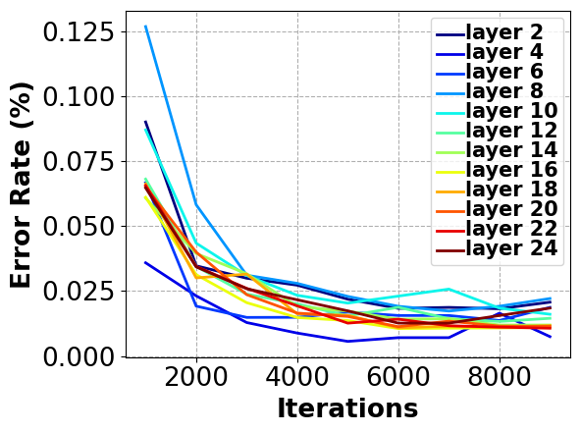}
    }
    \subfigure[Error rate per 1,000 iterations]{
        \label{fig:gpt3-350m-ARIMA-1k-bar}
        \includegraphics[width=0.35\linewidth]{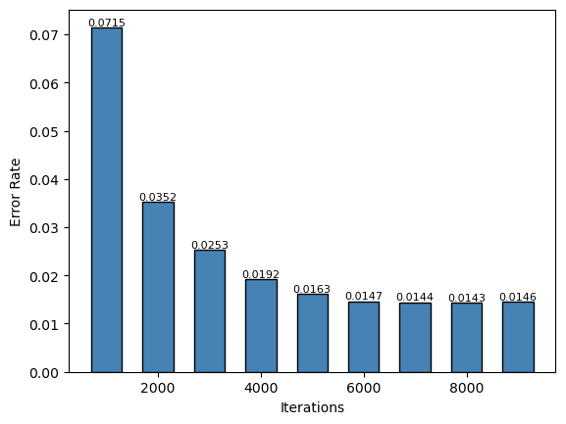}
    }
    \caption{Accuracy of ARIMA-based prediction for GPT-3 350M}
    \label{fig:gpt3-350m-ARIMA}
\end{figure}
\section{Evaluation}\label{sec:evaluation}
\subsection{\textit{Experiment setup 1}: GPT-3 125M}
In \textit{Experimental Setup 1}, the prediction algorithms are intended to predict the load proportions of all experts in each MoE layer for the next 1,000 iterations. Compared to the real data, we calculate the mean value of the error ratio for each individual MoE layer prediction result in sliding way. 
\begin{itemize}
    \item LSTM-based prediction. We use the expert load data obtained from two independent training as the training set and test set respectively, and the results are shown in Fig. \ref{fig:gpt3-125m-LSTM-1k}. 
    
    The transient and stable states of model training can be observed from the changing trend of prediction error, i.e., the prediction accuracy in the transient state (at the early stage of model training) gradually improves with the training iterations, and drops to less than $1\%$ and stays relatively stable after reaching the stable state (after training about 5,000 iterations indicated by the experimental result).
    
    \item ARIMA-based prediction. The experimental parameters are $ARIMA(5, 1, 5)$, which achieves a lower error rate than the above LSTM-based algorithm. As shown in the Fig. \ref{fig:gpt3-125m-ARIMA-1k}, the prediction results can be close to $0.5\%$ error rate at the stable state for each MoE layer. 
    
    \item SW\_Avg-based prediction. Although this method is computationally simple, it performs best among the three algorithms. In the stable state (i.e., after 5,000 iterations training), the algorithm predicts the expert load for each MoE with an error rate of about $0.25\%$, as demonstrated in the Fig. \ref{fig:gpt3-125m-AVG-1k}, which can provide more valuable guidance for resource allocation.
\end{itemize}

\subsection{\textit{Experiment setup 2}: GPT-3 350M}
\subsubsection{LSTM-based prediction}
Similarly, in \textit{Experimental Setup 2}, the prediction error rate results of the LSTM-based prediction algorithm are shown in Fig. \ref{fig:gpt3-350m-LSTM-1k}. Moreover, to exhibit the guidance value of the prediction method for resource allocation more directly, we give the discrete prediction results with the granularity of per 1,000 iterations instead of the sliding calculating way, as shown in the Fig. \ref{fig:gpt3-350m-LSTM-1k-bar}, which can be observed that the prediction error rates approximate less than $10\%$ at every 1,000 iterations from the second prediction.

\subsubsection{ARIMA-based prediction}
Similarly, we conducted the same experiments and evaluations for the ARIMA-based forecasting method, which demonstrates better performance in terms of prediction accuracy and stability compared to the LSTM-based prediction. Once the model training reaches the stable state, as shown in the Fig. \ref{fig:gpt3-350m-ARIMA-1k-bar}, the error rate of this algorithm stabilizes at approximately $1.4\%$.

\subsubsection{SW\_Avg-based prediction}
\begin{figure}[t]
    \centering
        \subfigure[Error rate]{
        \label{fig:gpt3-350m-AVG-1k}
        \includegraphics[width=0.35\linewidth]{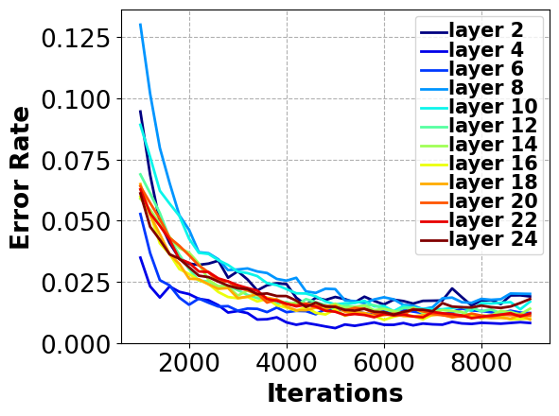}
    }
    \subfigure[Error rate per 1,000 iterations]{
        \label{fig:gpt3-350m-AVG-1k-bar}
        \includegraphics[width=0.35\linewidth]{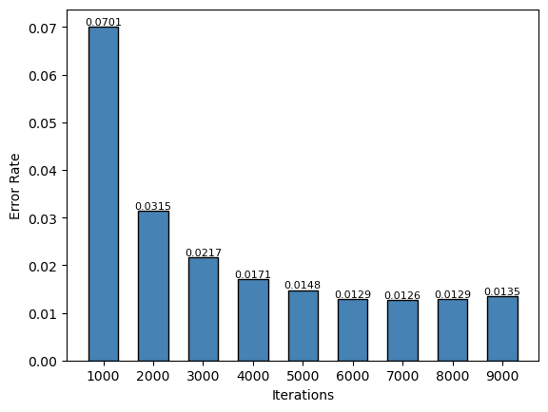}
    }
    \caption{Accuracy of SW\_Avg-based prediction for next 1,000 iterations}
    \label{fig:gpt3-350m-AVG-1k}
\end{figure}

\begin{figure}[t]
    \centering
        \subfigure[Error rate]{
        \label{fig:gpt3-350m-AVG-2k}
        \includegraphics[width=0.35\linewidth]{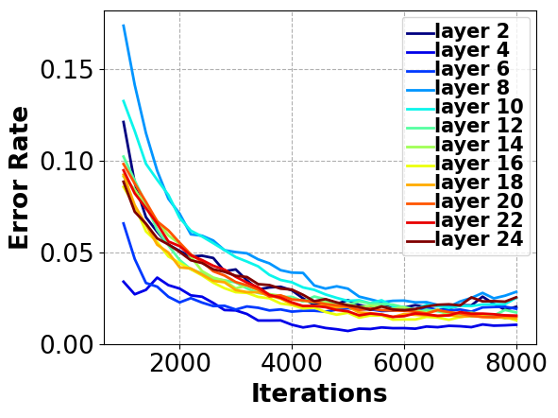}
    }
    \subfigure[Error rate per 2,000 iterations]{
        \label{fig:gpt3-350m-AVG-2k-bar}
        \includegraphics[width=0.35\linewidth]{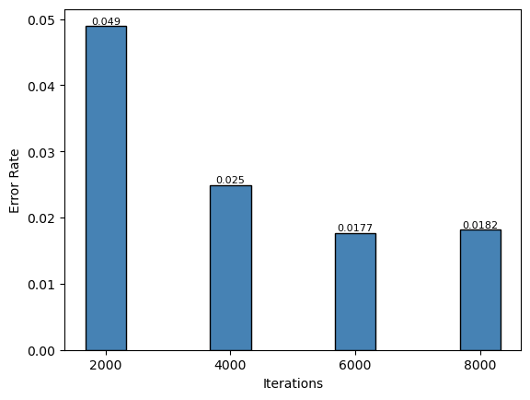}
    }
    \caption{Accuracy of SW\_Avg-based prediction for next 2,000 iterations}
    \label{fig:fig:gpt3-350m-AVG-2k}
\end{figure}

Consistent with \textit{Experimental Setup 1}, SW\_Avg prediction performs better than the other two algorithms, as exhibited in the Fig. \ref{fig:gpt3-350m-AVG-1k}, whose prediction error rate stabilizes at about $1.3\%$ in the stable state. 
In addition, we also conduct experiments to evaluate the prediction algorithm with the step size of 2,000 iterations. As presented in the Fig. \ref{fig:gpt3-350m-AVG-2k}, the accuracy trend of prediction results is consistent with aforementioned results, which stabilizes at about $1.7\%$ in the stable state.

%% file: conclusion.tex
\section{Conclusion}\label{sec:conclusion}
The emergence of MoE makes the computational complexity no longer scale up linearly with the volume of model parameters, which facilitates the development of LLM. The key idea of MoE is to use a gate network to assign tokens to selected experts for processing, which may result in a non-equilibrium load on processing tokens among the experts in each MoE layer during the model training. To this end, in this paper, we analyze the load variation characteristics of experts by conducting extensive experiments and correspondingly define the transient and stable states of model training. Moreover, we deploy three classical prediction algorithms based on the expert load states and achieve high-precision expert load prediction, which can provide valuable guidance on resource allocation for model training. 

\textit{In progress}: We are investigating the transient and stable states prediction algorithms based on this paper, and designing sensible and subtle resource allocation schemes to optimize the MoE architecture-based large-scale model training.

%% file: appendix.tex
\clearpage
\section*{Appendix}\label{sec:appendix}
\begin{figure*}[h!p]
    \centering
    \subfigure[Experts in layer-2]{
        \label{subfig:gpt3-350m-layer2-variance}
        \includegraphics[width=0.45\textwidth]{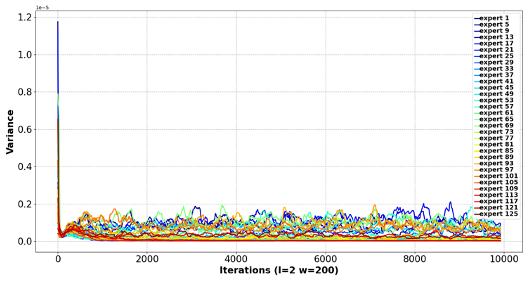}
    }
    \subfigure[Experts in layer-4]{
        \label{subfig:gpt3-350m-layer4-variance}
        \includegraphics[width=0.45\textwidth]{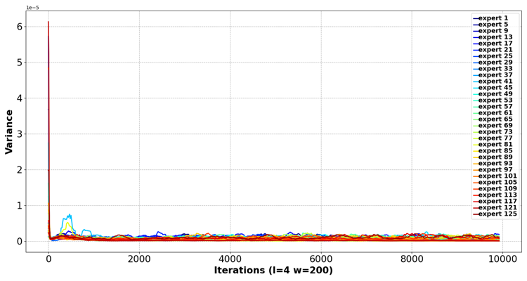}
    }
    \vspace{-0.2cm}
    \subfigure[Experts in layer-6]{
        \label{subfig:gpt3-350m-layer6-variance}
        \includegraphics[width=0.45\textwidth]{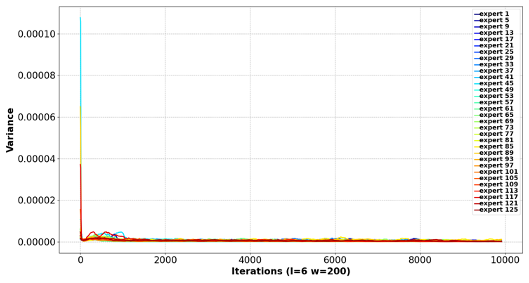}
    }
    \subfigure[Experts in layer-8]{
        \label{subfig:gpt3-350m-layer8-variance}
        \includegraphics[width=0.45\textwidth]{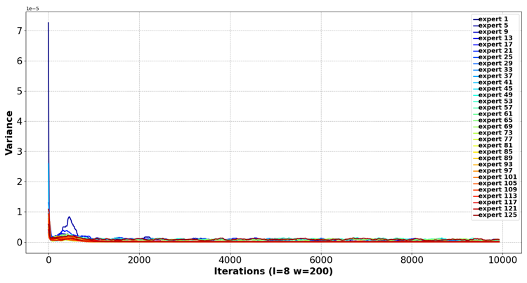}
    }
    \subfigure[Experts in layer-12]{
        \label{subfig:gpt3-350m-layer12-variance}
        \includegraphics[width=0.45\textwidth]{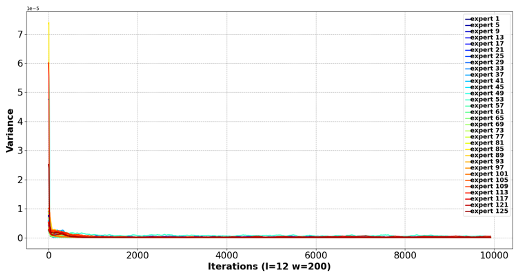}
    }
    \subfigure[Experts in layer-16]{
        \label{subfig:gpt3-350m-layer16-variance}
        \includegraphics[width=0.45\textwidth]{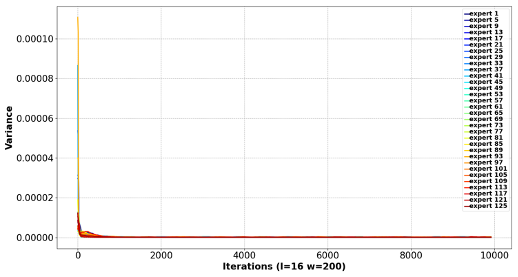}
    }
    \vspace{-1cm}
    \subfigure[Experts in layer-20]{
        \label{subfig:gpt3-350m-layer20-variance}
        \includegraphics[width=0.45\textwidth]{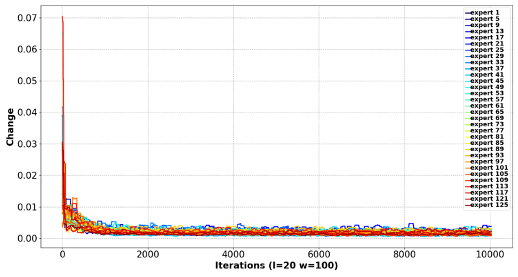}
    }
    \subfigure[Experts in layer-24]{
        \label{subfig:gpt3-350m-layer12-variance}
        \includegraphics[width=0.45\textwidth]{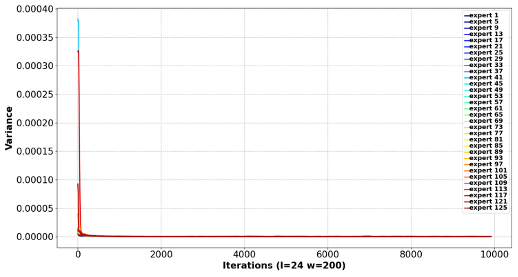}
    }
    \caption{\textsl{Variance} values of experts load proportion of GPT-3 350M (\textit{w=200})}
    \label{fig:gpt-3-350m-variance}
\end{figure*}

\begin{figure*}[t]
    \centering
    \subfigure[Experts in layer-2]{
        \label{subfig:gpt3-350m-layer2-range}
        \includegraphics[width=0.45\textwidth]{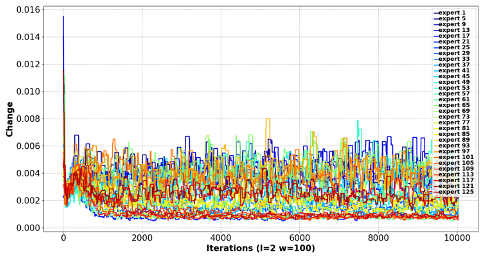}
    }
    \subfigure[Experts in layer-4]{
        \label{subfig:gpt3-350m-layer4-range}
        \includegraphics[width=0.45\textwidth]{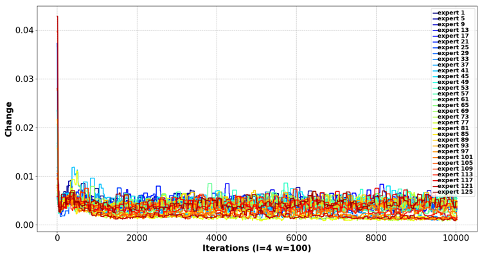}
    }
    \subfigure[Experts in layer-6]{
        \label{subfig:gpt3-350m-layer6-range}
        \includegraphics[width=0.45\textwidth]{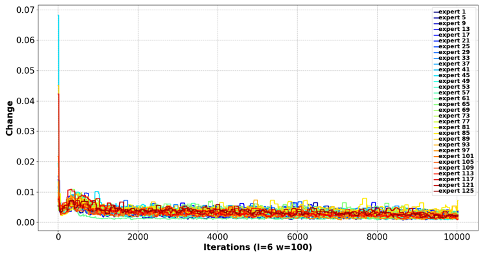}
    }
    \subfigure[Experts in layer-8]{
        \label{subfig:gpt3-350m-layer8-range}
        \includegraphics[width=0.45\textwidth]{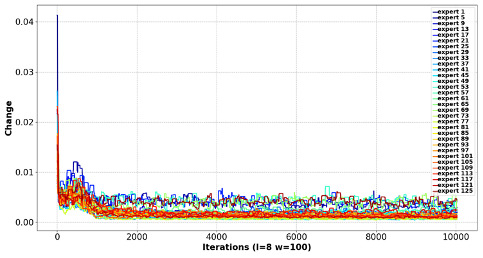}
    }
    \caption{\textsl{Range} values of experts load proportion of GPT-3 350M (\textit{w=200})}
    \label{fig:gpt-3-350m-range}
\end{figure*}

We analyzed the expert loads of the GPT3 350M model during training, and the \textit{variance} and \textit{range} of the each expert load proportion in each MoE layer at a sliding window of 100 are shown in Fig. \ref{fig:gpt-3-350m-variance} and Fig. \ref{fig:gpt-3-350m-range}. In general, the variation of expert load in Experimental Setup 2 also satisfies the defined transient and stable states.

